\documentclass[runningheads]{llncs}
\usepackage{graphicx}

\usepackage{tikz}
\usepackage{comment} 
\usepackage{amsmath,amssymb} 
\usepackage{color}

\usepackage{capt-of}

\usepackage{subfigure}
\usepackage{array}
\usepackage{booktabs}

\usepackage[marginal]{footmisc}

{\raise0.5ex\hbox{$#1$}\! \left/ \! \lower0.5ex\hbox{$#2$}\right.}

\def\ie{\emph{i.e.,~}}
\def\eg{\emph{e.g.,~}}

\def\etal{{\em et al.}}

\def\ToI{{\textit{ToI}~}}
\def\IrT{{\textit{IrT}~}}

\begin{document}
\pagestyle{headings}
\mainmatter
\def\ECCVSubNumber{100}  

\title{Adversarial Learning for Zero-shot Domain Adaptation} 

\titlerunning{Adversarial Learning for Zero-shot Domain Adaptation}
%
\author{Jinghua Wang\orcidID{0000-0002-2629-1198} \and
Jianmin Jiang\orcidID{0000-0002-7576-3999}}
\authorrunning{Wang J. and Jiang J.}
%
\institute{Research Institute for Future Media Computing, College of Computer Science \& Software Engineering, and Guangdong Laboratory of Artificial Intelligence \& Digital Economy (SZ), Shenzhen University, Shenzhen, China.
	\\
\email{\{wang.jh,jianmin.jiang\}@szu.edu.cn}\footnote{The corresponding author: Jianmin Jiang}}
\maketitle

\begin{abstract}
Zero-shot domain adaptation (ZSDA) is a category of domain adaptation problems where neither data sample nor label is available for parameter learning in the target domain.
With the hypothesis that the shift between a given pair of domains is shared across tasks,
we propose a new method for ZSDA by transferring domain shift from an \textit{irrelevant task} (\textit{IrT}) to the \textit{task of interest} (\textit{ToI}).
Specifically, we first identify an \textit{IrT}, where dual-domain samples are available,  and capture the domain shift with a coupled generative adversarial networks (CoGAN) in this task.
Then, we train a CoGAN for the \textit{ToI} and restrict it to carry the same domain shift as the CoGAN for \textit{IrT} does. 
In addition, we introduce a pair of co-training classifiers to regularize the training procedure of CoGAN in the \textit{ToI}.
The proposed method not only derives machine learning models for the non-available target-domain data, but also synthesizes the data themselves. 
We evaluate the proposed method on benchmark datasets and achieve the state-of-the-art performances. 
\keywords{Transfer Learning; Domain Adaptation; Zero-shot Learning; Coupled Generative Adversarial Networks}
\end{abstract}

\section{Introduction}

When a standard machine learning technique learns a model with the training data and applies the model on the testing data, it implicitly assumes that the testing data share the distribution with the training data \cite{alexNet,sanet,sumproduct}. 
However, this assumption is often violated  in applications, as the data in real-world are often from different domains \cite{Torralba-cvpr2011-unbiased}.
For example, the images captured by different cameras follow different distributions due to the variations of resolutions, illuminations, and capturing views.

Domain adaptation techniques tackle  the problem of domain shift by transferring knowledge from the label-rich source domain to the label-scarce target domain \cite{Ganin-icml15-unsupervised,Kodirov-iccv2015-unsupervisedDA-ZSL,ChenL0H19}. 
They have a wide range of applications, such as   
person re-identification \cite{Zhong_2019_CVPR},
semantic segmentation \cite{Luo_2019_CVPR}, 
attribute analysis \cite{Zhu-learning-classifier-icml2019},
and medical image analysis \cite{GhassamiKHZ18-neurips-2018}.
Most domain adaptation techniques assume that the data in target domain are available at the training time for model learning \cite{Yao2015Semi-supervisedDA,Lopez2012Semi,Ganin-icml15-unsupervised,Pinheiro_2018_CVPR}.
However, this is not always the case in the real-world.
For example, we may want a artificial intelligence system to provide continuous service with a newly installed camera \cite{Li2018Learning}.
This involves a domain adaptation task, where the source domain consists of the images captured by the old camera and the target domain consists of the non-accessible images captured by the new camera.
Such a task is referred to as domain generalization \cite{Ghifary2015Domain} or zero-shot domain adaptation (ZSDA) \cite{Peng-2018-eccv-zero-shot,Wang_2019_ICCV}. 

In this paper, we propose a new method to tackle the challenging ZSDA tasks, where only the source-domain data is available in the \textit{Task of Interest (ToI)}.
It is recognized that the existence of domain shift does not allow us to learn a model for the target domain based on the source-domain data alone.
To solve this problem, we establish a hypothesis that the domain shift, which intrinsically characterizes the difference between the domains, is shared by different tasks. 
For successful ZSDA, we firstly learn the domain shift from an \textit{irrelevant task (IrT)}  where many data in both domains are available, 
then transfer this domain shift to the \ToI and learn the model for the target domain.

\begin{figure}[!t]
	\begin{center}
		\includegraphics[width=0.65\linewidth]{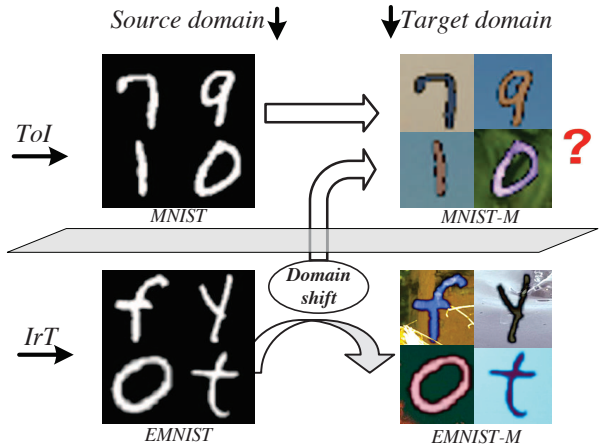}
	\end{center}
	\vspace{-18pt}
	\caption{An intuitive example of ZSDA (best viewed in color).
		The \ToI is digit image analysis and the \IrT is letter image analysis.
		The source domain consists of gray scale images and the target domain consists of color images.
		In order to learn the model for  the unseen \textit{MNIST-M} (\ie target-domain data in \textit{ToI}), we first learn the domain shift based on the dual-domain samples in the  \textit{IrT}, then transfer it to the \textit{ToI}.
		%
	}
	\label{fig:intuition}
\end{figure}

We illustrate an example of ZSDA in Fig. \ref{fig:intuition},  which learns a model for the color digit images (\ie \textit{MNIST-M} \cite{Ganin-icml15-unsupervised}), given the grayscale digit images (\ie \textit{MNIST} \cite{mnist}), the grayscale letter images (\ie \textit{EMNIST} \cite{emnist}), and the color letter images (\ie \textit{EMNIST-M}). 
In this example, the \ToI and \IrT are  digit and letter image analysis, respectively. 
The source domain consists of grayscale images, and the target domain consists of color images. 
We consider these two tasks to have the same domain shift, which transforms grayscale images to color images.

With the available dual-domain data in the  \textit{IrT}, we can train a coupled generative adversarial networks (CoGAN) \cite{Liu-NIPS2016_CoGan} (\ie \textit{CoGAN-IrT}) to model the joint distribution of images in these two domains.
This \textit{CoGAN-IrT} not only shows the sharing of two domains in high-level concepts, 
but also implicitly encodes the difference between them, which is typically referred to as domain shift. 
We consider one source-domain sample and one target-domain to be \textit{paired samples}  if they are realizations of the same thing and correspond to each other.
Fig. \ref{fig:intuition} shows eight grayscale images and their correspondences in the color domain. 
The RGB image and depth image of the same scene are also paired samples \cite{eccv}.
Based on the observation that it is the domain shift that introduces the difference between \textit{paired samples}, we define the domain shift to be the distribution of representation difference between \textit{paired samples}.

For successful ZSDA in the \textit{ToI}, we train a \textit{CoGAN-ToI} to capture the joint distribution of dual-domain samples and use it to synthesize the unseen samples in the target domain.
Besides the available samples in the source domain, we introduce two supervisory signals for \textit{CoGAN-ToI} training.
Firstly, we transfer the domain shift from \IrT to \ToI and enforce the \textit{CoGAN-ToI} to encode the same domain shift with \textit{CoGAN-IrT}. 
In other words, we restrict that the representation difference between paired samples to follow the same distribution in two tasks.
To improve the quality of the synthesized target-domain samples, we also take a pair of co-training classifiers to guide the training procedure of \textit{CoGAN-ToI}.
The predictions of these two classifiers are trained to be (i) consistent when receiving samples from both \IrT and \textit{ToI}, and (ii) different as much as possible when receiving samples which are not from these two tasks.
In the training procedure, we guide the \textit{CoGAN-ToI} to synthesize such target-domain samples that the classifiers produce consistent predictions when taking them as the input.
With domain shift preservation and co-training classifiers consistency as the supervisory signals, our \textit{CoGAN-ToI} can synthesize high quality data for the non-accessible target domain and learn well-performed models.

To summary, we propose a new method for ZSDA by learning across both domains and tasks and our contributions  can be highlighted in two folds.
\begin{itemize}
	\item Firstly, we propose a new strategy for domain adaptation through domain shift transferring across tasks.
	For the first time, we define the domain shift to be the distribution of representation difference between paired samples in two domains.
	We learn the domain shift from a \textit{CoGAN-IrT} that captures the joint distribution of dual-domain samples in the \IrT and design a method for shift transferring to the \textit{ToI}, where only source domain is seen.
	In addition to domain shift preservation, we also take the consistency of two co-training classifiers as another supervisory signal for  \textit{CoGAN-IoT} training to better explore the non-accessible target domain.
	\item Secondly, our method has a broader range of applications than existing methods \cite{Peng-2018-eccv-zero-shot,Wang_2019_ICCV}. 
	While our method is applicable when paired samples in the \IrT are non-accessible,   the work \cite{Peng-2018-eccv-zero-shot} is not.
	While our method can learn the domain shift from one \IrT and transfer it to multiple different \textit{ToI}s, the work \cite{Wang_2019_ICCV} is only applicable to a given pair of \textit{(IrT, ToI)}.
\end{itemize}

\section{Related Work}

While standard machine learning methods involves with a single domain \cite{resnet-HeZRS16,accv/WangJ18}, domain adaptation uses labeled data samples in one or more source domains to learn a model for the target domain.
For transferable knowledge learning, researchers normally minimize the discrepancy between domains by learning domain-invariant features \cite{Ganin-icml15-unsupervised,Long-nips-2016,Tzeng2014Deep,Yan-cvpr2017-mind-the-class}.
Ganin and Lempitsky   \cite{Ganin-icml15-unsupervised} introduced  gradient reversal layer to extract features that can confuses the domain classifier.
Long \etal  \cite{Long-nips-2016} introduced residual transfer network to bridge the source domain and the target domain by transferable features and adaptive classifiers.
Taking maximum mean discrepancy (MMD) as the measurement between domains, Tzeng \etal  \cite{Tzeng2014Deep} introduced an adaptation layer to learn representations which are not only domain invariant but also semantically meaningful.
In order to solve the problem of class weight bias, Yan \etal  \cite{Yan-cvpr2017-mind-the-class} introduced weighted MMD and proposed a classification EM algorithm for unsupervised domain adaptation.

These methods achieve good performances in various computer vision tasks.
However, none of them can solve the ZSDA problem as they rely on the target-domain data at the training time.
%
The existing techniques for ZSDA can be summarized into three categories based on their strategies.

The first strategy learns domain-invariant features which not only work in the available source domains but also generalize well to the unseen target domain. 
Domain-invariant component analysis (DICA) \cite{Muandet-Domain-icml2013} is a kernel-based method that learns a common feature space for different domains while preserving the posterior probability.
For cross domain object recognition, multi-task autoencoder (MTAE) \cite{Ghifary2015Domain} extends the standard denoising autoencoder framework by reconstructing the analogs of a given image for all domains. 
Conditional invariant deep domain generalization (CIDDG) \cite{Li_2018_ECCV}  introduces an invariant adversarial network to align the conditional distributions across domains and guarantee the domain-invariance property.
With a structured low-rank constraint, deep domain generalization framework (DDG) \cite{ding-fu-tip-deepDomainGeneralization} aligns multiple domain-specific networks to learn sharing knowledge across source domains.

The second strategy assumes that a domain is jointly determined by a sharing latent common factor and a domain specific factor 
\cite{Khosla2012Undoing,Li2017Deeper,Yang-zero-shot-Domain}. 
This strategy identifies the  common factor through decomposition and expects it to generalize well in the unseen target domain.
Khosla \etal \cite{Khosla2012Undoing} model each dataset as a biased observation of the visual world and conduct the decomposition via max-margin learning.
Li \etal \cite{Li2017Deeper} develop a low-rank parameterized CNN model to simultaneously exploit the relationship among domains and learn the domain agnostic classifier.
Yang and Hospedales \cite{Yang-zero-shot-Domain} parametrise the domains with continuous values and propose a solution to predict the subspace of the target domain via manifold-valued data regression.
Researchers also correlate domains with semantic descriptors \cite{Kodirov-iccv2015-unsupervisedDA-ZSL} or latent domain vectors \cite{Kumagai-2018arxiv-ZeroshotDA}.

The third strategy first learns the correlation between domains from an assistant task, then accomplishes ZSDA based on the available source-domain data and the domain correlation \cite{Peng-2018-eccv-zero-shot,Wang_2019_ICCV}.
Normally, this strategy relies on  an \IrT where data from both source and target domain  are sufficiently available.
In comparison with the first two strategies, this strategy can work well with a single source domain.
Zero-shot deep domain adaptation (ZDDA) \cite{Peng-2018-eccv-zero-shot} aligns representations from source domain and target domain in the \IrT and expect the alignment in the \textit{ToI}.
CoCoGAN \cite{Wang_2019_ICCV} aligns the representation across tasks in the source domain and takes the alignment as the supervisory signal in the target domain.

\section{Background} 

\textbf{Generative Adversarial Networks} (GAN) consists of two competing models, \ie the generator and the discriminator \cite{Goodfellow-nips2014-gan}.
Taking a random vector $ z \sim p_z $ as the input, the generator aims to synthesize images $ g(z) $ which are resemble to the real image as much as possible.
The discriminator tries to distinguish real images from the synthesized ones. It takes an image $ x $ as the input and outputs a scalar $ f(x) $, which is expected to be large for real images and small for synthesized images.
The following objective function formulates the adversarial training procedure of the generator and the discriminator:
\begin{equation}
\begin{aligned}
\begin{split}
\max\limits_{g}\min\limits_{f} V(f,g)  \equiv E_{x\sim p_x}[-\log f(x)]
+E_{z\sim p_z}[-\log (1-f(g(z)))],
\end{split}
\end{aligned}
\label{eq:gan-objective-function}
\end{equation}
where $ E $ is the empirical estimate of expected value of the probability.
In fact, Eq. (\ref{eq:gan-objective-function}) measures the Jensen-Shannon divergence between the distribution of real images and that of the synthesized images \cite{Goodfellow-nips2014-gan}.

\textbf{Coupled Generative Adversarial Networks} (CoGAN) consists of a pair of GANs (\ie GAN$ _1 $ and GAN$ _2 $) which are closely related with each other. 
With each GAN corresponds to a domain, CoGAN captures the joint distribution of images from two different domains \cite{Liu-NIPS2016_CoGan}.
Let $ x_i\sim p_{x_i} (i=1,2) $ be the images from the $ i $th domain.
In GAN$ _i (i=1,2) $, we denote the generator  as  $ g_i $ and the discriminator as $ f_i $.
Based on a sharing random vector $ z $, the generators synthesize image pairs $ (g_1(z),g_2(z)) $ which not only are indistinguishable from the real ones but also have correspondences.
We can formulate the objective function of the CoGAN as follows:
\begin{equation}
\small
\begin{split}
\begin{aligned}
\max\limits_{g_1,g_2}\min\limits_{f_1,f_2}V(f_1,f_2,g_1,g_2)\equiv
& E_{x_1\sim p_{x_1}}[-\log f_1(x_1)]
+E_{z\sim p_z}[-\log (1-f_1(g_1(z)))]
\\
+&E_{x_2\sim p_{x_2}}[-\log f_2(x_2)]
+E_{z\sim p_z}[-\log (1-f_2(g_2(z)))],
\end{aligned}
\end{split}
\end{equation}
subject to two constraints: 
(i) $\theta_{g_1^j}=\theta_{g_2^j}$, $1\leq j \leq n_{g}$; and (ii) 
$\theta_{f_1^{n_1-k}}=\theta_{f_2^{n_2-k}}$, $0\leq k\leq n_{fs}-1$. 
The parameter $ n_i (i=1,2) $ denotes the number of layers in the discriminator $ f_i $.
While the first constraint restricts the generators to have $ n_{g} $ sharing bottom layers,
the second restricts the discriminators to have $ n_{f} $ sharing top layers.
These two constraints force the generators and discriminators to process the high-level concepts in the same way,
so that the CoGAN is able to discover the correlation between two domains.
CoGAN is effective in dual-domain analysis, as it is capable to learn the joint distribution of data samples (\ie $p_{x_1,x_2} $) based on the  samples drawn individually from the marginal distributions (\ie $p_{x_1}$ and $p_{x_2}$).

\section{Approach}
\label{sec:approach}

\subsection{Problem Definition}

We define a domain $ D=\{X,P(X)\} $ to be the data sample space $ X  $ and its marginal probability distribution $ P(X) $ \cite{Pan-yang-a-survey-on-transferLearning}. 
Given the data samples $ X $, a task $ T=\{Y,P(Y|X)\} $ consists of a label space $ Y $ and the conditional probability distribution $ P(Y|X) $. 
This work considers two tasks to be the same as long as they have sharing label space.
In \textit{ToI}, the label space is $Y^{\alpha} $, the source domain is $ D_s^{\alpha}= \{X_s^{\alpha},P(X_s^{\alpha})\}$, and the target domain is $ D_t^{\alpha}= \{X_t^{\alpha},P(X_t^{\alpha})\}$.
Then, the \ToI is denoted by $ T^{\alpha}=\{Y^{\alpha},P_s(Y^{\alpha}|X_s^{\alpha})\} \cup \{Y^{\alpha},P_t(Y^{\alpha}|X_t^{\alpha})\} $.  

Given the labeled data samples in the source domain (\ie $ (x_s^{\alpha},y_s^{\alpha}) $, $ x_s^{\alpha} \in X_s^{\alpha} $ and $ y_s^{\alpha} \in Y^{\alpha} $),
our ZSDA task aims to derive the  conditional probability distribution $ P(Y^{\alpha}|X_t^{\alpha}) $ in the target domain.
In general, the main challenge of this task is induced by non-accessibility of the target domain, as well as the domain shift, \ie  $ P(X_s^{\alpha})\neq P(X_t^{\alpha}) $ and $P_s(Y^{\alpha}|X_s^{\alpha}) \neq P_t(Y^{\alpha}|X_t^{\alpha}) $.

\begin{figure*}[!t]
	\begin{center} 
		\includegraphics[width=\linewidth]{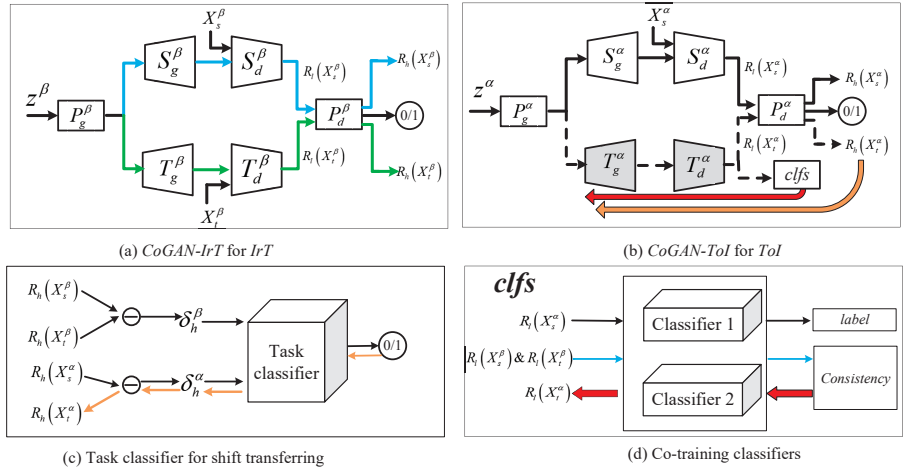}
	\end{center}
	\caption{
		The network structures of our method. 
		The \textit{CoGAN-IrT} in (a) models the joint distribution of $ (x_s^{\beta},x_t^{\beta}) $  in the \textit{IrT}. 
		The \textit{CoGAN-ToI} in (b) models the joint distribution of $ (x_s^{\alpha},x_t^{\alpha}) $  in the \textit{ToI}.
		In the discriminators of these two CoGANs, we use $ R_l(\cdot) $ to denote the lower level representation produced by the non-sharing layers, and $ R_h(\cdot) $ to denote higher level  representations produced by the sharing layers, respectively.
		The task classifier in (c) discriminates $ \delta_h^{\beta}=R_h(x_t^{\beta})\ominus R_h(x_s^{\beta}) $ from $ \delta_h^{\alpha}=R_h(x_t^{\alpha})\ominus R_h(x_s^{\alpha}) $.
		We maximize the loss of this task classifier to align the domain shift.
		The co-training classifiers in (d) produce the labels for $ X_s^{\alpha} $ and consistent predictions for $ X_s^{\beta} $ and $ X_t^{\beta} $.
		To train the \textit{CoGAN-ToI},
		we use domain shift preservation to regularize the higher level features and co-training classifiers to regularize the lower level features.
		The back-propagation directions of these two signals are marked by orange and red, respectively.
	}
	\label{fig:network}
\end{figure*}

\subsection{Main Idea}

In order to accomplish the ZSDA task, we identify an irrelevant task (\textit{IrT}) that satisfies two constraints: 
(i) the \IrT involves the same pair of domains with the \textit{ToI}; 
and (ii) the dual-domain samples in the \IrT are available.
Under the hypothesis that the shift between a given pair of domains maintains across tasks, we propose to learn  the domain shift from \IrT and transfer it to \textit{ToI}.

Let the label space of \IrT be  $ Y^{\beta} $.  
We denote the \IrT as $ T^{\beta}=\{Y^{\beta},P_s(Y^{\beta}|X_s^{\beta})\} \cup \{Y^{\beta},P_t(Y^{\beta}|X_t^{\beta})\} $, where   $ D_s^{\beta}= \{X_s^{\beta},P(X_s^{\beta})\}$ is the source domain   and $ D_t^{\beta}= \{X_t^{\beta},P(X_t^{\beta})\}$ is the target domain, respectively. 
Note that, the source-domain samples $ (X_s^{\alpha} $ and $ X_s^{\beta} $)  are in the same sample space.
It is also true in the target domain.
In the example of Fig. \ref{fig:intuition}, while the source-domain data $ X_s^{\alpha}=$\textit{MNIS}T and $ X_s^{\beta}=$\textit{EMNIST} are  grayscale images, the target-domain data $ X_t^{\alpha}=$\textit{MNIST-M} and $ X_t^{\beta}=$\textit{EMNIST-M} are color images.

In this work, we define two corresponding samples from source domain and target domain as \textit{paired samples}.
In most cases, two \textit{paired samples} are different views of the same object. 
For example, a grayscale image in \textit{MNIST} and its corresponding color image in \textit{MNIST-M} are paired samples in Fig. \ref{fig:intuition}.
The depth image and RGB image of the same scene are also \textit{paired samples}.
While the similarity between \textit{paired samples} is determined by the object itself, their difference is mainly introduced by the domain shift.
Our work only assumes the existences of correspondence between dual-domain samples.
Nevertheless, the correspondences between them in the \IrT are not required.

For correlation analysis between \textit{paired samples}, we train CoGANs to capture the joint distribution of source-domain and target-domain data.
As both $ X_s^{\beta} $ and $ X_t^{\beta} $ are available, we can easily train \textit{CoGAN-IrT} (Fig. \ref{fig:network} (a)) for the \IrT using the standard method \cite{Liu-NIPS2016_CoGan}. 
The main difficulty lies in the training of \textit{CoGAN-ToI} (Fig. \ref{fig:network} (b)) for the \textit{ToI}, as the target-domain data $ X_t^{\alpha} $ is not available.
To tackle this problem, we propose two kinds of supervisory information for  \textit{CoGAN-ToI} training, which are \textit{domain shift preservation} and \textit{co-training classifiers consistency}.

For easier transferring across tasks, we define domain shift to be the distribution of element-wise difference between \textit{paired samples} in the representation space.
We can learn the domain shift from the \textit{CoGAN-IrT} which carries the correlation between two domains by varying the inputting noise $ z^{\beta} $.
After that, we train \textit{CoGAN-ToI} and enforce the representation difference between \textit{paired samples} of \ToI to follow the distribution learned in \textit{CoGAN-IrT} by maximizing the loss of a task classifier.
Fig. \ref{fig:network} (c) visualizes the task classifier which aims to identify the task label of the representation difference.
In this way, the domain shift is transferred from the \textit{IrT} to \textit{ToI}.

To better explore the unseen target domain of \textit{ToI}, we also build two co-training classifiers (Fig. \ref{fig:network} (d)) and use their consistency to guide the training procedure of  \textit{CoGAN-ToI}.
By enforcing the weights of the classifiers to be different from each other as much as possible, we aim to analyze data samples from distinct views.
The classifiers are trained to: 
(i) predict the labels of  $ X_s^{\alpha} $;
(ii) produce consistent predictions when receiving  $ X_s^{\beta} $ and $ X_t^{\beta} $;
and (iii) produce different predictions when receiving samples not involved with \textit{ToI} or \textit{IrT}.
Thus, we can use the consistency of these two classifiers to evaluate whether a sample is involved with the two tasks.
We guide the training procedure of \textit{CoGAN-ToI} to synthesize $ X_t^{\alpha} $ as such that the classifiers also produce consistent predictions when receiving their representations.

\subsection{Training}

In \textit{CoGAN-IrT} (Fig. \ref{fig:network} (a) ), the sharing layers are $ P_g^{\beta} $ and $ P_d^{\beta} $, the non-sharing layers are $ S_g^{\beta} $, $ T_g^{\beta} $, $ S_d^{\beta} $ and $ T_d^{\beta} $. 
The components of \textit{CoGAN-ToI} are denoted in similar way in Fig. \ref{fig:network} (b).
For simplicity,  we use $ R_l(x) $ to denote the lower level representation of sample $ x $ produced by the non-sharing layers and 
and $ R_h(x) $ to denote the higher level representation  produced by sharing layers.
Note, the representation extraction procedures $ R_l(\cdot) $ and $ R_h(\cdot) $ vary with task and domain.

\textbf{Domain shift}

We train \textit{CoGAN-IrT} based on the dual-domain samples ($ X_s^{\beta}$ and $ X_t^{\beta} $)  and let it carry the correlation between two domains.
The \textit{CoGAN-IrT} can synthesize a set of \textit{paired samples} for the  \textit{IrT}. 
For two \textit{paired samples} $ (x_s^{\beta} \in X_s^{\beta},x_t^{\beta}  \in X_t^{\beta}) $, we characterize their shift by  the element-wise difference between representations in a sharing layer, \ie $ \delta_h^{\beta}=R_h(x_t^{\beta})\ominus R_h(x_s^{\beta})$.
We then define the domain shift to be the  distribution of $ \delta_h^{\beta} $, \ie $ p_{\delta_h^{\beta}} $.
Specifically, we can obtain a set of $ \{\delta_h^{\beta}\} $ by feeding \textit{CoGAN-IrT} with different values of inputting noise $ z^{\beta} $.

\textbf{Co-training classifiers}

Both of the two co-training classifiers (denoted as $ clf_1 $ and $ clf_2 $) take the representations (\ie $ R_l(X_s^{\beta}) $, $ R_l(X_t^{\beta}) $, and $ R_l(X_s^{\alpha}) $)  as the input.
With $ R_l(x) $ as the input, the classifier $ clf_i $ ($ i=1,2 $) produces a $ c $-dimensional vector $ v_i (x) $, where $ c $ denotes the number of categories in $ X_s^{\alpha} $.
We minimize the following loss to train the classifiers:
\begin{equation}
\small
L(clf_1,clf_2)=
\lambda_{w}L_{w} (w_1,w_2)
+
\lambda_{acc} L_{cls}(X_s^{\alpha})
-
\lambda_{con}L_{con}(X^{\beta})
+
\lambda_{diff}L_{diff}(\tilde{X}),
\label{eq:co-classifier-objective}
\end{equation}
where $ L_{w} $ measures the similarity between the two classifiers,
$ L_{cls} $ denotes the loss to classify the labeled source-domain samples of \textit{ToI}, 
$ L_{con} $ assesses the consistency of the output scores when receiving dual-domain samples of \textit{IrT} (\ie $ X_s^{\beta} $ and  $ X_t^{\beta} $) as the input,
and $ L_{diff} $ assess the consistency when receiving  samples $ \tilde{X} $ which are not related \textit{ToI} or \textit{IrT}.

As in standard co-training methods \cite{Saito-cvpr2018-maximum-classifier},
we expect the two classifiers to have diverse parameters so that they can analyze the inputs from different views.
In this work, we implement these two classifiers with the same neural network structure and assess their similarity by the cosine distance between the parameters:
\begin{equation}
\small
L_{w}=\left. w_1^T*w_2 \middle/ ||w_1||*||w_2|| \right. ,
\label{eq:weight-loss}
\end{equation}
where $ w_i $ is the vectored parameters of $ clf_i $.

With the labeled source-domain data in \textit{ToI}, we can easily formulate a multi-class classification problem and use the soft-max loss to define the second term of Eq. (\ref{eq:co-classifier-objective}) as follows:
\begin{equation}
\small
L_{cls}=  
-\sum_{x_s^{\alpha}\in X_s^{\alpha}}
\sum_{i=1}^{2}
\sum_{j=1}^c v_i^j(x_s^{\alpha}) l^j(x_s^{\alpha}),
\label{eq:acc-loss}
\end{equation}
where $ v_i^j(x_s^{\alpha}) $ is the $ j $th element of the prediction $  v_i(x_s^{\alpha}) $, 
the binary value $ l^j(x_s^{\alpha}) $ denotes whether $ x_s^{\alpha} $ belongs to the $ j $th class or not. 
This item regularizes the classifiers to produce semantically meaningful vectors.

Different from $ X_s^{\alpha} $ in \textit{ToI}, the labels for dual-domain data in \IrT are not available. 
It is impossible to predict their true labels.
To gain supervisory signals from these label-missing data, we restrict the two classifiers to produce consistent predictions.
The consistency for a given sample is measured by the dot product of its two predictions.
Thus, we define the third term in Eq. (\ref{eq:co-classifier-objective}) as:
\begin{equation}
\small
L_{con}=  
\sum_{x^{\beta}\in X_s^{\beta}\cup X_t^{\beta}}
v_1(x^{\beta})\cdot v_2(x^{\beta}).
\label{eq:con-loss}
\end{equation}

The last term $ L_{diff} $ regularizes the classifiers to produce different predictions when receiving samples that are not related with the two tasks.
It is defined in the same way to $ L_{con} $ and the only difference lies in the input $ \tilde{X} $.
Here, the samples in $ \tilde{X} $ have two sources: 
(i) the samples in public datasets, \eg imageNet \cite{imageNet}; 
(ii) the corrupted images by replacing a patch of $ x_s^{\beta} $, $ x_t^{\beta} $, and $ x_s^{\alpha} $ with random noise.

In principal, we can use the consistency of these two classifiers to assess whether a sample is properly involved with \textit{IrT} or \textit{ToI} in these two domains. 
Thus, we can guide the training procedure of  \textit{CoGAN-ToI} in such a way that the synthesized $ X_t^{\alpha} $ should satisfy $ v_1(X_t^{\alpha})= v_2(X_t^{\alpha})$, and take this as a complementary supervisory signal of domain shift preservation.

\textbf{CoGAN-ToI}

At this stage, we train \textit{CoGAN-ToI} to capture the joint distribution of \textit{paired samples} in the \textit{ToI}.
By correlating the two domains, a well-trained \textit{CoGAN-ToI} is able to synthesize the non-available target-domain data.
We use three constraints to train \textit{CoGAN-ToI}, including (i) one branch captures the distribution of $ X_s^{\alpha} $; (ii) the domain shift is shared by the two tasks, \ie $  p_{\delta_{h}^{\beta}} = p_{\delta_{h}^{\alpha}}$, where $ \delta_{h}^{\alpha}=R_h(x_t^{\alpha})\ominus R_h(x_s^{\alpha}) $; and (iii) the co-training classifiers have consistent predictions for the synthesized sample $ x_t^{\alpha} $, \ie $ v_1(x_t^{\alpha})=v_2(x_t^{\alpha}) $.

This work trains the two branches of \textit{CoGAN-ToI} separately, 
unlike the standard method in \cite{Liu-NIPS2016_CoGan} that trains them simultaneously.
To satisfy the first constraint, we consider the source-domain branch (consisting of $ P_g^{\alpha}$, $S_g^{\alpha}$, $S_d^{\alpha}$, and $P_d^{\alpha}$) as an independent GAN and train it using  the available  $ X_s^{\alpha} $.

Though involving in different tasks, both $ X_t^{\beta} $ and $ X_t^{\alpha} $ are images from the target domain.
Thus, they are composed of the same set of low-level details.
In order to mimic the processing method learned in the \textit{IrT}, we initialize the non-sharing components of \textit{CoGAN-ToI} in the target domain as  $T_g^{\beta} \rightarrow T_g^{\alpha} $  and $T_d^{\beta} \rightarrow T_d^{\alpha} $.

After initialization, we use the second and third constraints to train the non-sharing components  ($ T_g^{\alpha} $ and $ T_d^{\alpha} $) for the target domain and fine-tune the sharing components ($ P_g^{\alpha} $ and $ P_d^{\alpha} $).
Specifically, we minimize the following loss function:
\begin{equation}
\footnotesize
\begin{aligned}
V(P_g^{\alpha},T_g^{\alpha},T_d^{\alpha},P_d^{\alpha})
\equiv  
\lambda_{con}^{\alpha}
\sum_{x_t^{\alpha}=g_t^{\alpha}(z^{\alpha})}
v_1(x_t^{\alpha})\cdot v_2(x_t^{\alpha})
-L_{clf}(\delta_{h}^{\alpha},\delta_{h}^{\beta})
,
\end{aligned}
\label{eq:adapt-nonsharing-objective-function}
\end{equation}
where $ g_t^{\alpha}=P_g^{\alpha}+T_g^{\alpha} $ is the generator.
While the first term assesses how the two classifiers agree with each other, the second term assesses how $ \delta_{h}^{\alpha} $ is distinguishable from $ \delta_{h}^{\beta} $.

With \textit{CoGAN-ToI}, we train a classifier for the synthesized target-domain data by three steps.
Firstly, we train a classifier $ \varPhi_s (\cdot) $ for the labeled source-domain data.
Then, we synthesize a set of paired samples $ (x_s^{\alpha},x_t^{\alpha}) $ and use $ \varPhi_s (\cdot) $ to predict their labels.
Finally, we train a classifier $ \varPhi_t  (\cdot)$ for $ x_t^{\alpha} $ with the constraint $ \varPhi_s (x_s^{\alpha})=\varPhi_t (x_t^{\alpha})$ and evaluate our method using the average accuracy.

\section{Experiments}
\label{sec:experiment}

\subsection{Adaptation Across Synthetic Domains}
\label{subsec:exp-synthetic}

We conduct experiments on four gray image datasets, including MNIST ($ D_M $) \cite{mnist}, Fashion-MNIST ($ D_F $) \cite{fashion-mnist}, NIST ($ D_N $) \cite{nist}, and EMNIST ($ D_E $) \cite{emnist}.  
Both MNIST \cite{mnist} and Fashion-MNIST have $ 70000 $ images from $ 10 $ classes. 
NIST is imbalance and has more than $ 40 $k images from $ 52 $ classes.
EMNIST has more than $ 145 $k images from $ 26 $ classes.

These four datasets are in the gray domain ($ G\textendash dom $). 
We create three more domains for evaluation,  
\ie the colored domain ($ C\textendash dom $), the edge domain ($ E\textendash dom $), and the negative domain ($ N\textendash dom $). 
The $ C\textendash dom $ is created  using the method in \cite{Ganin-icml15-unsupervised}, \ie combining an image with a random color patch in BSDS500 \cite{bsds500-amfm_pami2011}. 
We apply canny detector to create   $ E\textendash dom $ and the operation of $ I_n=255-I $ to create   $ N\textendash dom $.

\textbf{Implementation details}

%
In order to learn transferable domain shift across tasks, the two CoGANs (\ie \textit{CoGAN-IrT} and  \textit{CoGAN-ToI}) have the same network structure.
The two branches inside these CoGANs also share the same structure, and both generators and discriminators have seven layers.
We transform the output of the last convolutional layer of discriminator into a column vector before feeding it into a single sigmoid function.
The last two layers in generators and the first two layers in discriminators are non-sharing layers for low-level feature processing.

The task classifier has four convolutional layers to identify the task label of its input.
We vary the input noise  $ z^{\beta} $ of  \textit{CoGAN-IrT} to extract $ R_h(x) $ and thus obtain a set of $ \delta_{h}^{\beta} $. 
The parameters of the task classifier are initialized with zero-centered normal distribution. 
We adopt the stochastic gradient descent (SGD) method for optimization. 
The batch size is set to be $ 128 $ and the learning rate is set to be $ 0.0002 $.

The  co-training classifiers are implemented as convolutional neural networks with three fully connected layers, with $ 200 $, $ 50 $, and $ c $, respectively.
We use  $ c $ to denote the number of categories in $ X_s^{\alpha} $.
We set the hyper-parameter as $ \lambda_{w}=0.01 $, $ \lambda_{acc}=1.0 $, $ \lambda_{con}=0.5 $, and
$ \lambda_{diff}=0.5 $.

The source-domain branch in  \textit{CoGAN-ToI} is firstly trained independently using the available data  $ X_s^{\alpha} $. 
Based on the two supervisory signals, we use back-propagation method to train $ T_g^{\alpha} $ and $ T_d^{\alpha} $. 
Simultaneously, the $ P_g^{\alpha} $ and $ P_d^{\alpha} $ are fine-tuned.
In our experiment, we train the two branches of \textit{CoGAN-ToI} in an iterative manner to obtain the best results.

\textbf{Results}

With the above four datasets, we conduct experiments on ten different pairs of (\textit{IrT, ToI}). 
Note,  $ D_N $ and $ D_E $ are the same task, as both of them consist of letter images. 
We test four pairs of source domain and target domain, including $ (G\textendash dom, C\textendash dom) $, $ (G\textendash dom, E\textendash dom) $, $ (C\textendash dom, G\textendash dom) $, and $ (N\textendash dom, G\textendash dom) $.

We take two existing methods as the benchmarks, including ZDDA \cite{Peng-2018-eccv-zero-shot} and CoCoGAN \cite{Wang_2019_ICCV}.
In addition, we adapt ZDDA by introducing a domain classifier in order to learn from non-corresponding samples and denote it as ZDDA$_{dc}$.
We also conduct ablation study by creating the baseline  \textit{CTCC}, which only uses \underline{C}o-\underline{T}raining \underline{C}lassifiers \underline{C}onsistency to train \textit{CoGAN-ToI}.

\newcommand{\tabincell}[2]{\begin{tabular}{@{}#1@{}}#2\end{tabular}}

\newcolumntype{P}[1]{>{\centering\arraybackslash}p{#1}}

\begin{table*}[htb]
	\centering
	\caption{The accuracy of different methods with $(source, target)= (G\textendash dom, C\textendash dom) $}
	\begin{tabular}{l
			|P{0.8cm}P{0.8cm}P{0.8cm}
			|P{0.8cm}P{0.8cm}P{0.8cm}
			|P{0.8cm}P{0.8cm}
			|P{0.8cm}P{0.8cm}
		}
		\hline 
		\textit{ToI} 
		& \multicolumn{3}{c|}{$D_M$} 
		& 
		\multicolumn{3}{c|}{$D_F$}
		& 
		\multicolumn{2}{c|}{$D_N$} 
		& 
		\multicolumn{2}{c}{$ D_E$} 
		\\ 		
		\cline{2-11}
		\textit{IrT} 
		&$D_F$& $D_N$ & $D_E$ 
		& $D_M$&$D_N$ &$D_E$ 
		& $ D_M $ & $D_F$
		& $ D_M $ & $D_F$ 
		\\ 	
		\hline 
		\hline
		ZDDA 
		&73.2 & 92.0  & 94.8
		&51.6 & 43.9 & 65.3 
		&34.3 & 21.9 
		&71.2 &  47.0
		\\ 
		CoCoGAN 
		&78.1  &  92.4  & \textbf{95.6}
		&56.8  &  56.7  & \textbf{66.8} 
		&41.0  &  44.9
		&\textbf{75.0}  &  54.8  
		\\ 
		ZDDA$_{dc}$  
		& 69.3 & 79.6 & 80.7 
		& 50.6 & 42.4 & 62.0 
		& 29.1 & 20.2 
		& 49.8 & 46.5
		\\
		CTCC  
		& 68.5 & 74.9 & 77.6 
		& 42.0 & 52.9 & 60.9 
		& 37.0 & 43.6 
		& 47.3 & 45.2		
		\\ 
		Ours  
		&\textbf{81.2}&\textbf{93.3}&95.0
		&\textbf{57.4}&\textbf{58.7}&62.0
		&\textbf{44.6}&\textbf{45.5}
		&72.4&\textbf{58.9}
		\\ 
		\hline
	\end{tabular}
	\label{tab:G-2-C}
\end{table*}

\begin{table*}[!htb]
	\centering
	\caption{The accuracy of different methods with $(source, target)= (G\textendash dom, E\textendash dom) $}
	\begin{tabular}{l
			|P{0.8cm}P{0.8cm}P{0.8cm}
			|P{0.8cm}P{0.8cm}P{0.8cm}
			|P{0.8cm}P{0.8cm}
			|P{0.8cm}P{0.8cm}
		}
		\hline 
		\textit{ToI} & \multicolumn{3}{c|}{$ D_M $} 
		& \multicolumn{3}{c|}{$  D_F$} 
		& 
		\multicolumn{2}{c|}{$ D_N $} 
		& 
		\multicolumn{2}{c}{$ D_E $} 
		\\ 		
		\cline{2-11}
		\textit{IrT} 
		&$D_F$& $D_N$ & $D_E$ 
		& $D_M$&$D_N$ &$D_E$ 
		& $ D_M $ & $D_F$
		& $ D_M $ & $D_F$ 
		\\ 	
		\hline \hline
		ZDDA & 
		72.5  & 91.5  &  93.2 & 
		54.1  & 54.0  &  65.8 & 
		42.3  & 28.4  &  
		73.6  & 50.7   
		\\  
		CoCoGAN & 
		79.6  & \textbf{94.9}  &  95.4 &   
		61.5  & 57.5  &  71.0 & 
		48.0  & 36.3  &  
		77.9  & 58.6
		\\
		ZDDA$_{dc}$ &  
		66.5 & 83.3 & 84.7 & 
		49.3 & 50.4 & 58.0 & 
		42.2 & 31.6 & 
		65.0 & 41.2
		\\ 
		CTCC &  
		65.5 & 73.9 & 80.5 & 
		44.0 & 40.8 & 37.3 & 
		40.0 & 31.4 & 
		57.7 & 48.2
		\\	 
		Ours 
		&\textbf{81.4}&93.5&\textbf{96.3}
		&\textbf{63.2}&\textbf{58.7}&\textbf{72.4}
		&\textbf{49.9}&\textbf{38.6}
		&\textbf{78.2}&\textbf{61.1}
		\\ 
		\hline
	\end{tabular}
	\label{tab:G-2-E}
\end{table*}

As seen in Tab. \ref{tab:G-2-C}-\ref{tab:N-2-G}, our method achieves the best performance in average. 
Taking $ D_E $ classification as an example, our method outperforms ZDDA \cite{Peng-2018-eccv-zero-shot} by a margin of $ 8.9\% $, and outperforms CoCoGAN \cite{Wang_2019_ICCV} by a margin of $ 4.1\% $ when \textit{IrT} is   $ D_F $ in Tab. \ref{tab:G-2-C}.
In average, our method performs $ 7.38\% $ better than ZDDA and $ 0.69\% $ better than CoCoGAN in Tab. \ref{tab:G-2-C}.

\begin{table*}[!h]
	\centering
	\caption{The accuracy of different methods with $(source, target)= (C\textendash dom, G\textendash dom) $}
	\begin{tabular}{l
			|P{0.8cm}P{0.8cm}P{0.8cm}
			|P{0.8cm}P{0.8cm}P{0.8cm}
			|P{0.8cm}P{0.8cm}
			|P{0.8cm}P{0.8cm}
		}
		\hline
		\textit{ToI} & \multicolumn{3}{c|}{$ D_M $} 
		& \multicolumn{3}{c|}{$  D_F$} 
		& 
		\multicolumn{2}{c|}{$ D_N $} 
		& 
		\multicolumn{2}{c}{$ D_E $} 
		\\ 		
		\cline{2-11}
		\textit{IrT} 
		&$D_F$& $D_N$ & $D_E$ 
		& $D_M$&$D_N$ &$D_E$ 
		& $ D_M $ & $D_F$
		& $ D_M $ & $D_F$ 
		\\ 	
		\hline \hline
		ZDDA  & 
		67.4  &  85.7 &  87.6  &  
		55.1  &  49.2  &  59.5 &  
		39.6  &  23.7  &  
		75.5  &   52.0   
		\\  
		CoCoGAN & 
		73.2 & 89.6 &  \textbf{94.7} &  
		61.1 & 50.7 &  70.2 & 
		47.5 & 57.7 & 
		80.2 & 67.4
		\\	
		ZDDA$_{dc}$ &  
		61.5 & 76.7 & 79.9 & 
		51.2 & 46.1 & 53.4 & 
		31.3 & 20.4 & 
		61.2 & 42.2 
		\\ 	
		CTCC &  
		62.1 & 76.9 & 68.6 & 
		47.2 & 45.6 & 57.6 & 
		27.5 & 33.6 & 
		58.0 & 49.9 
		\\	
		Ours
		&\textbf{73.7}&\textbf{91.0}&93.4
		&\textbf{62.4}&\textbf{53.5}&\textbf{71.5}
		&\textbf{50.6}&\textbf{58.1}
		&\textbf{83.5}&\textbf{70.9}
		\\ 	
		\hline
	\end{tabular}
	\label{tab:C-2-G}
\end{table*}

\begin{table*}[!h]
	\centering
	\caption{The accuracy of different methods with $(source, target)= (N\textendash dom, G\textendash dom) $}
	\begin{tabular}{l
			|P{0.8cm}P{0.8cm}P{0.8cm}
			|P{0.8cm}P{0.8cm}P{0.8cm}
			|P{0.8cm}P{0.8cm}
			|P{0.8cm}P{0.8cm}
		}
		\hline
		\textit{ToI} & \multicolumn{3}{c|}{$ D_M $} 
		&
		\multicolumn{3}{c|}{$  D_F$} 
		& 
		\multicolumn{2}{c|}{$ D_N $} 
		& 
		\multicolumn{2}{c}{$ D_E $} 
		\\ 		
		\cline{2-11}
		\textit{IrT} 
		&$D_F$& $D_N$ & $D_E$ 
		& $D_M$&$D_N$ &$D_E$ 
		& $ D_M $ & $D_F$
		& $ D_M $ & $D_F$ 
		\\ 	
		\hline \hline
		ZDDA  & 
		78.5 & 90.7 & 87.6  &  
		56.6 & 57.1 & 67.1 & 
		34.1 & 39.5 & 
		67.7 & 45.5     
		\\ 		
		CoCoGAN & 
		80.1  & 92.8  & 93.6  & 
		63.4  & 61.0  & 72.8 & 
		47.0  & 43.9  & 
		78.8  & 58.4 
		\\ 
		ZDDA$_{dc}$ & 
		68.4 & 79.8 & 82.5 & 
		48.1 & 46.2 & 64.6 & 
		28.6 & 34.4 & 
		61.8 & 36.2
		\\ 
		CTCC & 
		68.4 & 80.0 & 80.2 & 
		50.1 & 55.1 & 61.3 & 
		37.6 & 33.9 & 
		56.1 & 33.9
		\\ 
		Ours 
		&\textbf{82.6}&\textbf{94.6}&\textbf{95.8}
		&\textbf{67.0}&\textbf{68.2}&\textbf{77.9}
		&\textbf{51.1}&\textbf{44.2}
		&\textbf{79.7}&\textbf{62.2}
		\\ 
		\hline
	\end{tabular}
	\label{tab:N-2-G}
\end{table*}

\begin{figure}[h]
	\centering
	\begin{minipage}[t]{0.4\textwidth}
		\centering
		\includegraphics[scale=0.22]{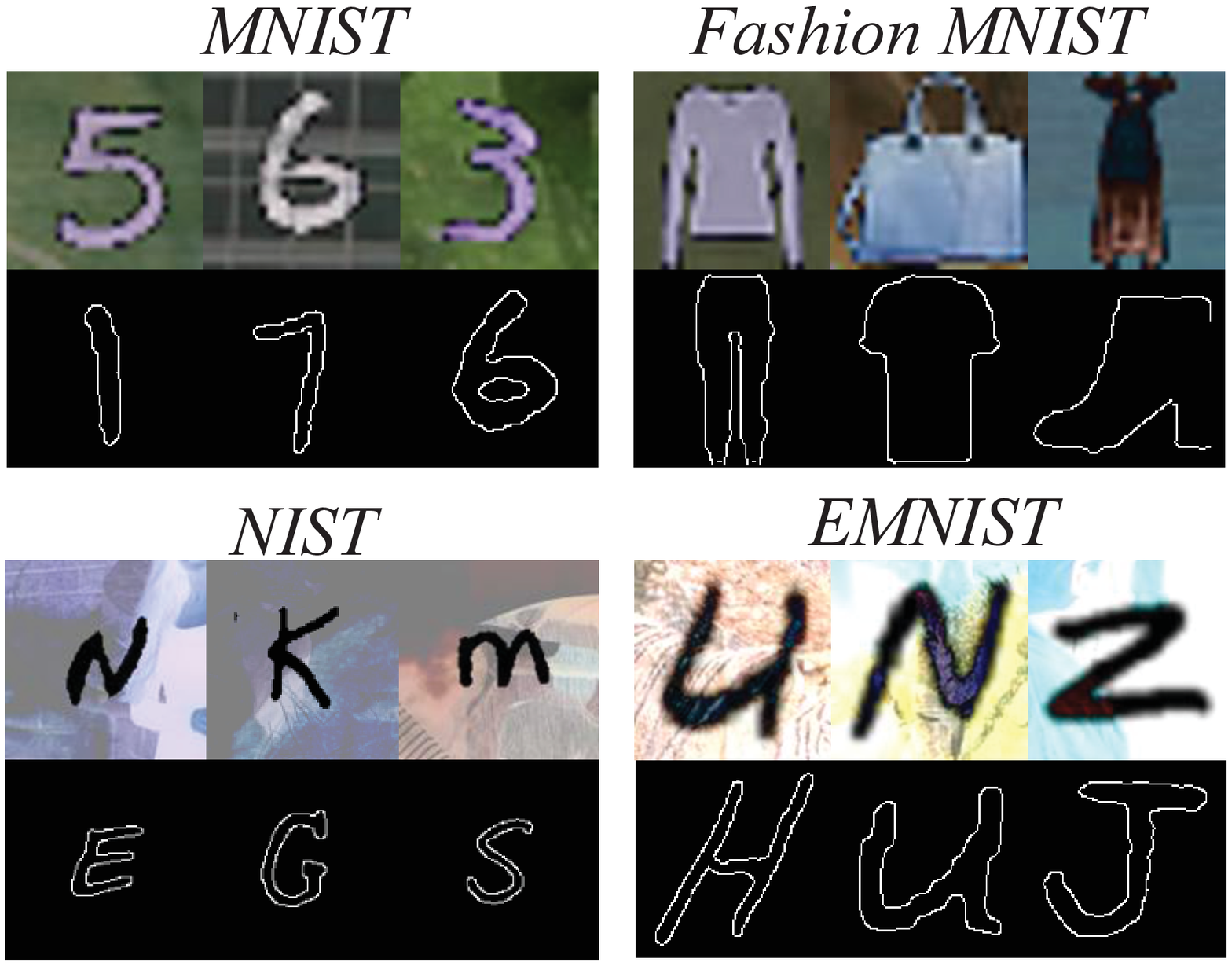}
		\caption{The generated images in the $ C\textendash dom $ and $ E\textendash dom $.}
		\label{fig:generatedImages}
	\end{minipage}
	\hfill
	\begin{minipage}[t]{0.55\textwidth}
		\centering
		\includegraphics[width=0.9\linewidth]{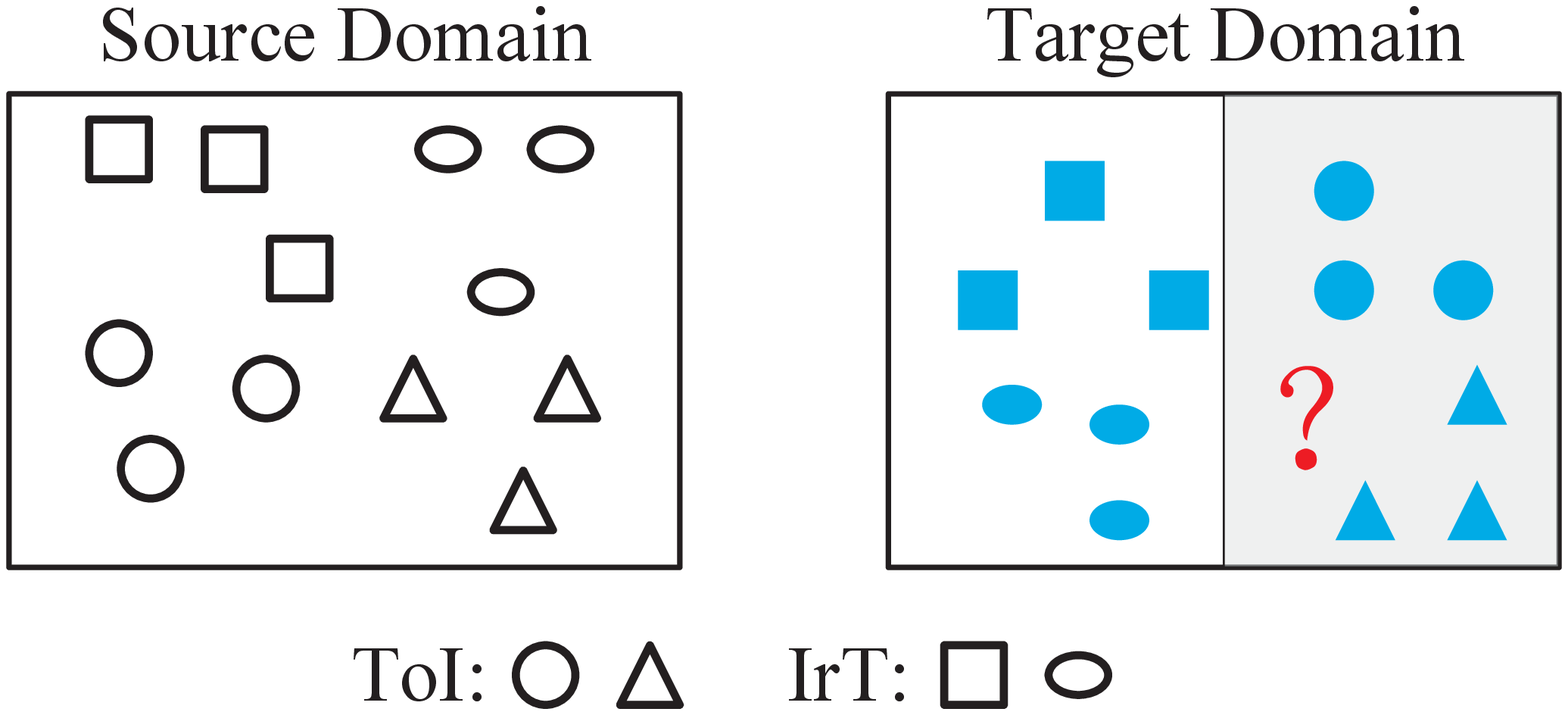}
		\captionof{figure}{An example. \textit{ToI} represents a subset of the categories (\textit{square} and \textit{triangle}) and \textit{IrT} represents the rest (\textit{circle} and \textit{ellipse}).}
		\label{fig:category-across-intuitive}
	\end{minipage}
\end{figure}

In each of the Tab. \ref{tab:G-2-C}-\ref{tab:N-2-G}, the proposed method improves CTCC more than $ 10\% $ in average.
This means that the domain shift transferring are useful in the training procedure of \textit{CoGAN-ToI}. 
Among the three tasks, the digit image classification is the easiest one.  
Out of all settings, the most successful one transfers knowledge from  the $ G \textendash dom $ to the $ E \textendash dom $ with  $ D_E $ as the \textit{IrT} and $ D_M $ as the \textit{ToI}.  
In this case, our method achieves the accuracy of $ 96.3\% $.
For $ D_M$ classification in $ G\textendash dom $, our method achieve the accuracy of $ 95.8\% $ with $ D_E $ as \IrT and $ N\textendash dom $ as the source domain, 
outperforming other techniques (including $ 89.5\% $ in  \cite{Haeusser-ICCV2017-associative}, and  $ 94.2\% $ in \cite{SaitoUH17-icml2017}) which rely on the availability of the target-domain data in the training stage.
With \textit{CoGAN-ToI}, we not only derive models for the unseen target domain,  but also synthesize data themselves.
Fig. \ref{fig:generatedImages} visualizes the generated images in  $ C\textendash dom $ and   $ E\textendash dom $ with $ G\textendash dom $ as the source domain.

%

%

%

\subsection{Adaptation in Public Dataset}

We also evaluate our method on  Office-Home   \cite{VenkateswaraECP17-office-home}, which has four different domains, \ie Art (\textbf{Ar}), Clipart (\textbf{Cl}), Product (\textbf{Pr}), and Real-world (\textbf{Rw}).
It has more than 15k images from $ 65 $ categories.

As it is difficult to identify an analogous set for this dataset, we evaluate our method on adaptation across subsets.
Give a pair of domains,
we take a subset of the categories as the \ToI and the rest as the \textit{IrT}. 
An example is shown Fig. \ref{fig:category-across-intuitive}, where the \ToI represents the classification of two categories (\textit{square} and \textit{triangle}), and \IrT represents the classification of other two categories (\textit{circle} and \textit{ellipse}).
Here, we set the parameters as $\lambda_w=0.01$, $\lambda_{acc}=1$, and $\lambda_{con}=\lambda_{diff}=0.1$.

\begin{table*}[!htb]
	\small
	\centering
	\caption{The accuracy of different methods on Office-Home}
	\begin{tabular}{l
			|P{0.65cm}P{0.65cm}P{0.65cm}
			|P{0.65cm}P{0.65cm}P{0.65cm}
			|P{0.65cm}P{0.65cm}P{0.65cm}
			|P{0.65cm}P{0.65cm}P{0.65cm}
		}
		\hline
		Source& \multicolumn{3}{c|}{\textit{Ar}} & \multicolumn{3}{c|}{\textit{Cl}} & \multicolumn{3}{c|}{\textit{Pr}} & \multicolumn{3}{c}{\textit{Rw}} 
		\\
		\hline
		Target& \textit{Cl}     & \textit{Pr}    & \textit{Rw}    & \textit{Ar}     & \textit{Pr}    & \textit{Rw}    & \textit{Ar}     & \textit{Cl}    & \textit{Rw}    & \textit{Ar}     & \textit{Cl}    & \textit{Pr}    
		\\
		\hline
		ZDDA$_{dc}$    & 
		53.2   & 61.4  & 68.8  &
		67.4   & 57.0  & 68.4  &
		60.9   & 40.6  & 62.4  & 68.1   & 43.4  & 50.3  
		\\
		CoCoGAN & 
		62.2   & 69.5  & 74.5  &
		66.7   & 74.0  & 66.4  &
		57.6   & 53.4  & 71.7  &
		69.2   & 51.3  & 65.8   
		\\
		CTCC    & 
		55.7   & 61.5  & 66.5  &
		66.8   & 64.6  & 65.2  &
		56.3   & 46.6  & 61.6  &
		64.3   & 43.7  & 57.7 
		\\
		Ours    &  \textbf{62.7}&\textbf{71.9}&\textbf{76.3}&
		\textbf{72.6}&\textbf{75.1}&\textbf{73.9}&
		\textbf{70.3}&\textbf{60.8}&\textbf{74.8}&
		\textbf{72.2}&\textbf{61.4}&\textbf{72.2}
		\\
		\hline
	\end{tabular}
	\label{tab:office-home}
\end{table*}

Let $ N_{\alpha} $ denote the number of categories of \textit{ToI}.
We fix the value of $ N_{\alpha} $ to be $ 10 $ and conduct experiments on all of the $ 12 $ possible different pairs of source domain and target domain.
As seen in Tab. \ref{tab:office-home}, our method achieves the best performances in all cases.
This indicates that our method is applicable to a broad range of applications.
Our method can beats both ZDDA and CoCoGAN by a margin larger than $ 10\% $, when source domain is \textit{Rw} and target domain is \textit{Cl}.

\begin{table}[!htb]
	\centering
	\small
	\caption{The variation of accuracy against parameter $\lambda_{con}^\alpha$}
	\begin{tabular}{l|
			P{1.2cm}P{1.2cm}P{1.2cm}
			P{1.2cm}P{1.2cm}P{1.2cm}
		}
		\hline
		$\lambda_{con}^\alpha$ & ar$\rightarrow$ cl & ar$\rightarrow$ pr & ar$\rightarrow$ rw & cl$\rightarrow$ ar & cl$\rightarrow$ pr & cl$\rightarrow$ rw
		\\ 
		\hline
		0.001                  
		& 59.3  & 68.5  & 73.3  
		& 65.7  & 68.3  & 69.3  
		\\
		0.005                  
		& 61.6  & 70.3  & 75.7  
		& 70.6  & 74.6  & 71.1  
		\\
		0.01                   
		& 62.7  & 71.9  & 76.3  
		& 72.6  & 75.1  & 73.9  
		\\
		0.02                   
		& 62.1  & 71.0  & 74.7  
		& 72.1  & 76.1  & 72.8  
		\\
		0.1                    
		& 53.0  & 64.8  & 66.1  
		& 60.4  & 60.4  & 63.5
		\\
		\hline
	\end{tabular}
	\label{tab:accuracy-with-different-parameters}
\end{table}

We use the parameter $ \lambda_{con}^{\alpha}= 0.01 $  to balance the two terms in Eq. (\ref{eq:adapt-nonsharing-objective-function}).
Generally, the CTCC mainly regularizes the training of $T_s^\alpha$, which processes the low-level details.
The detail-richer $X_t^\beta$ means more knowledge are available for training, and the more transferable across tasks the $T_s^\alpha$ is.
Thus, we set smaller value for $\lambda_{con}^\alpha$ when richer details are included in $X_t^\beta$.
Tab. \ref{tab:accuracy-with-different-parameters} lists the accuracy of our method on Office-Home with different parameter values of $\lambda_{con}^\alpha$.
As seen, our method performs well when $\lambda_{con}^\alpha \in [0.005,0.01]$.

Let $ N_s $ be the number of samples in the $X=\{X_s^\alpha,X_s^\beta,X_t^\beta\}$.
We use $ 2N_s $ supplementary samples to train the $ L_{diff} $ where (i) half are randomly cropped from the ImageNet and (ii) half are obtained by replacing patches of training samples with random noises.
Tab. \ref{tab:num-supplementary} lists the performance of our method with different number of supplementary samples.
As seen, $ 2N_s $ supplementary samples are enough for model training.

\begin{table}[!htb]
	\small
	\centering
	\caption{The variation of accuracy against number of supplementary samples}
	\begin{tabular}{l|
			P{1.2cm}P{1.2cm}P{1.2cm}
			P{1.2cm}P{1.2cm}P{1.2cm}
		}
		\hline
		Num & 
		ar$\rightarrow$ cl & ar$\rightarrow$ pr & ar$\rightarrow$ rw & cl$\rightarrow$ ar & cl$\rightarrow$ pr & cl$\rightarrow$ rw
		\\ 
		\hline
		0.8N 
   		& 60.3  & 67.5  & 73.4  
		& 68.8  & 67.0  & 70.7  
		\\
		N    
		& 61.3  & 70.7  & 73.6  
		& 70.3  & 73.2  & 71.0  
		\\
		1.6N 
		& 62.5  & 71.5  & 76.0  
		& 71.5  & 74.3  & 73.5  
		\\
		2N   
		& 62.7  & 71.9  & 76.3  
		& 72.6  & 75.1  & 73.9  
		\\
		4N   
		& 62.7  & 71.9  & 76.3  
		& 72.7  & 75.1  & 73.9
		\\
		\hline
	\end{tabular}
	\label{tab:num-supplementary}
\end{table}

\section{Conclusion and Future Work}

This paper proposes a new method for ZSDA based on the hypothesis that different tasks may share the domain shift for the given two domains. We learn the domain shift from one task and transfer it to the other by bridging two CoGANs with a task classifier. Our method takes the domain shift as the distribution of the representation difference between paired samples and transfers it across CoGANs. Our method is capable of not only learning the machine learning models for the unseen target domain, but also generate target-domain data samples. Experimental results on six datasets show the effectiveness of our method in transferring knowledge among images in different domains and tasks.

The proposed method learns the shift between domains and transfers it across tasks.  This strategy makes our method to be applicable only when ``large” shift exists across domains, such as (rgb, gray), (clipart, art) etc. Thus, our method cannot perform well on the datasets where the domain shift is “small”, such as VLSC and Office-31.
In the future, we will train a classifier to determine whether correspondence exists between a source-domain sample and a synthesized target-domain sample.
Such a classifier can guide the training procedure of CoGAN, even when only samples from a single domain is available.

\section*{Acknowledgment}

The authors wish to acknowledge the financial support from: (i) Natural Science Foundation China (NSFC) under the Grant no. 61620106008 ; (ii) Natural Science Foundation China (NSFC) under the Grant no. 61802266.

%
%
\bibliographystyle{splncs04}
\bibliography{DomainShiftTransfer}

\end{document}